\documentclass{article}
\usepackage{iclr2026_conference,times}
\iclrfinalcopy
\PassOptionsToPackage{numbers,compress,sort}{natbib}

\usepackage{times}
\usepackage{latexsym}
\usepackage{makecell}
\usepackage[T1]{fontenc}
\usepackage[utf8]{inputenc}
\usepackage{microtype}
\usepackage{graphicx}
\usepackage{amsmath}
\usepackage{amsmath,amssymb,amsthm}
\usepackage{hyperref}
\usepackage{url}
\usepackage{booktabs}
\usepackage{amsfonts}
\usepackage{xcolor,colortbl}
\usepackage{multirow}
\usepackage{xspace}
\usepackage{pifont}

\usepackage{caption}
\usepackage{subcaption}
\usepackage{amsthm}

\definecolor{mygrey}{HTML}{cfcfcf}
\definecolor{mygreen}{HTML}{8de5a1}
\definecolor{myblue}{HTML}{a1c9f4}
\definecolor{mypurple}{HTML}{d0bbff}
\definecolor{myred}{rgb}{1.0, 0.6, 0.6}
\definecolor{darkgreen}{HTML}{006401}
\definecolor{darkred}{HTML}{8B0000}
\definecolor{reasoningllm}{RGB}{251,221,203}
\definecolor{basellm}{RGB}{222,244,252}
\definecolor{reasoningllm2}{RGB}{255,200,165}

\newcommand{\symbols}[1]{\texttt{\small{#1}}}

\newcommand{\cmark}{\textcolor{green}{\ding{51}}}
\newcommand{\xmark}{\textcolor{red}{\ding{55}}}

\newcommand{\benchmark}{\textsc{SLR-Bench}\xspace}

\title{LLMs Gaming Verifiers: \\ RLVR can Lead to Reward Hacking}
\author{
Lukas Helff\textsuperscript{1,2,3},\;
Quentin Delfosse\textsuperscript{1,4},\;
David Steinmann\textsuperscript{1,2},\;
Ruben Härle\textsuperscript{1,5},\;
Hikaru Shindo\textsuperscript{1},\\
\textbf{
Patrick Schramowski\textsuperscript{1,2,3,6},\;
Wolfgang Stammer\textsuperscript{7},\;
Kristian Kersting\textsuperscript{1,2,3,5},\;
Felix Friedrich\textsuperscript{8}\thanks{\small work done while at TU Darmstadt}}\\
$^{1}$TU Darmstadt\quad 
$^{2}$hessian.AI\quad 
$^{3}$DFKI\quad 
$^{4}$Intrinsic\quad 
$^{5}$Lab1141\\
$^{6}$CERTAIN, Germany\quad
$^{7}$MPI-Inf, SIC\quad
$^{8}$Meta FAIR\\
}

\begin{document}
\maketitle

\begin{abstract}
As reinforcement Learning with Verifiable Rewards (RLVR) has become the dominant paradigm for scaling reasoning capabilities in LLMs, a new failure mode emerges: \textit{LLMs gaming verifiers}. We study this phenomenon on inductive reasoning tasks, where models must induce and output logical rules. We find that RLVR-trained models systematically abandon rule induction. Instead of learning generalizable patterns (e.g., ``trains carrying red cars go east''), they enumerate instance-level labels, producing outputs that pass verifiers without capturing the relational patterns required by the task.
We show that this behavior is not a failure of understanding but a form of reward hacking: imperfect verifiers that check only extensional correctness admit false positives. To detect such shortcuts, we introduce \textit{Isomorphic Perturbation Testing} (IPT), which evaluates a single model output under both extensional and isomorphic verification, where the latter enforces invariance under logically isomorphic tasks. While genuine rule induction remains invariant, shortcut strategies fail.
We find that shortcut behavior is specific to RLVR-trained reasoning models (e.g., GPT-5, Olmo3) and absent in non-RLVR models (e.g., GPT-4o, GPT-4.5, Ministral). Moreover, shortcut prevalence increases with task complexity and inference-time compute. In controlled training experiments, extensional verification directly induces shortcut strategies, while isomorphic verification eliminates them. These results show that RLVR can incentivize reward hacking not only through overt manipulation but also by exploiting what the verifier fails to enforce.\looseness-1
\end{abstract}

\section{Introduction}
Reinforcement learning (RL) has become the dominant paradigm for scaling reasoning capabilities, powering frontier models like OpenAI’s GPT-5 and GPT-5-mini. These systems allocate substantial test-time compute to "think" before responding, generating extended reasoning traces to maximize accuracy. While this approach has driven impressive performance on complex mathematical and logical benchmarks~\citep{openai_o3_system_card_2025}, it introduces a fundamental tension. When reward signals rely on imperfect proxies, models learn to exploit the evaluation mechanism instead of solving the intended task~\citep{baker2025monitoring}. This has manifested as explicit reward hacking: models overwrite unit tests, monkey-patch scoring functions, delete assertions, or force early program termination to obtain a passing score without implementing the correct solution~\citep{krakovna2020specification, skalse2022defining, macdiarmid2025emergent, metr2025, zhong2025impossiblebench}.

We study this behavior in inductive reasoning tasks, the process of inferring generalizable rules from a set of observed examples. For instance, after observing alien plants where \symbols{plant\_01} has purple leaves and is toxic, and \symbols{plant\_02} has green leaves and is safe, an inductive reasoner should induce a rule such as “Plants with purple leaves are toxic.” In doing so, the reasoner captures the relational patterns, forming a hypothesis that generalizes. Upon encountering a new plant, \symbols{plant\_03}, with purple leaves, the reasoner predicts toxicity without direct observation.

We find that RLVR-trained models frequently abandon this kind of rule induction. Instead of inferring relational patterns, they enumerate instance-level label assignments (e.g., plant\_01 is toxic, plant\_02 is safe''). These outputs are semantically vacuous with respect to the task's objective, yet reflect a precise strategy. A verifier that checks only extensional consistency (e.g., whether \symbols{plant\_01} is toxic and \symbols{plant\_02} is safe) yields false positives despite the absence of inductive reasoning. We term this behavior a \textit{reward shortcut}: the model exploits implicit assumptions in what the verifier treats as correct. \looseness=-1

To diagnose this behavior, we introduce \textit{Isomorphic Perturbation Testing} (IPT), which evaluates a model's output under two regimes: extensional verification on the original task, and isomorphic verification on a logically isomorphic perturbation obtained by permuting object identifiers while preserving relational structure. Since genuine rule induction is invariant under such transformations while extensional enumerations are not, a shortcut is identified whenever an output passes extensional but fails isomorphic verification. This provides a black-box criterion for detecting shortcut reliance in frontier models where weights, activations, and reasoning traces are inaccessible.
Across our evaluation, we find that RLVR-trained models (GPT-5 family, Olmo3) exhibit systematic shortcut behavior, while non-RLVR models (GPT-4o, GPT-4.5, Ministral) exhibit none on identical tasks. Shortcut prevalence increases with both task complexity and inference-time compute, suggesting that additional compute may be directed toward exploiting verifier weaknesses rather than improving generalization. We train two identical models using Olmo-3's RLVR pipeline~\cite{olmo2025olmo3}, differing only in the verifier used for reward. Purely extensional verification directly induces a growing hacking gap between extensional and isomorphic reward, while isomorphic verification eliminates it.\looseness=-1

Overall, we contribute: (1) evidence of systematic reward shortcuts in RLVR-trained models on inductive reasoning tasks, absent in non-RLVR models; (2) Isomorphic Perturbation Testing, a black-box method for detecting shortcuts in closed-source models; (3) analysis linking shortcut prevalence to task complexity and inference-time compute; and (4) evidence that extensional verification induces reward hacking, while isomorphic verification prevents it. \looseness=-1

\section{Related Work}
\textbf{Reward Hacking.} Reward hacking in reinforcement learning refers to agents exploiting weaknesses in reward specifications rather than solving the intended task~\citep{krakovna2020specification}. As RL has scaled to LLMs, analogous behaviors have emerged in increasingly complex environments~\citep{macdiarmid2025emergent,wang2026implicit}. In agentic and coding settings, RL-trained models manipulate evaluation mechanisms by overwriting unit tests, monkey-patching scoring functions, deleting assertions, or prematurely terminating programs to obtain passing scores without producing correct solutions~\citep{metr2025,baker2025monitoring}. These failures are commonly described as environmental hacking, where agents interfere with external validation. Our work identifies a subtler failure mode in reasoning tasks: models exploit implicit assumptions on the verifier’s notion of correctness, producing outputs that satisfy proxy evaluation criteria while evading the intended reasoning objective.\looseness-1

\textbf{Inductive Logic Programming (ILP).} 
ILP studies the problem of learning a general hypothesis $H$ (a logic program) from background knowledge $B$ and labeled examples $(E^+, E^-)$ such that $B \wedge H$ entails all positive examples (completeness) while remaining consistent with the negative ones (consistency)~\citep{Cropper2021InductiveLP,ILP_Muggleton,DeRaedt2008}. ILP aims to generalize intensional patterns (rule-based representations) that can assign labels beyond extensional representations (explicit instance-level facts). While classical ILP focuses on algorithms for hypothesis search, we adopt this formal perspective as a diagnostic lens for assessing whether LLMs perform genuine rule induction or rely on extensional shortcut strategies.

\section{Isomorphic Perturbation Testing}\label{sec:setup}
How can we determine whether LLMs genuinely perform reasoning, rather than exploiting weaknesses in the evaluation protocol? This question is increasingly pressing as LLMs are optimized via RLVR, and imperfect rewards can incentivize misalignment and reward hacking~\citep{metr2025,zhong2025impossiblebench}. 
Detecting such shortcut behavior is especially challenging for frontier LLMs, since weights, activations, and reasoning traces are inaccessible, leaving evaluation limited to final outputs. To address this, we introduce \textit{Isomorphic Perturbation Testing} (IPT), a methodology for diagnosing shortcut behavior based solely on model outputs. IPT builds upon a simple logical principle: genuine inductive rule learning is invariant to logically isomorphic tasks.

\textbf{Setup.} To analyse shortcut behaviour we adopt \benchmark~\citep{helff2025slr}, which frames reasoning as a sequence of ILP tasks. In each task, the model is provided with background knowledge $B$ describing trains and their cars (e.g., car colors), along with labeled examples: eastbound (positive examples $E^+$) and westbound (negative examples $E^-$). The objective is to induce a hypothesis $H$ -- a minimal logic rule that explains the labeling by abstracting relational patterns from the background knowledge. For instance, a valid hypothesis could be: “A train is eastbound if it carries a red car.”

\textbf{From Induction to Reward Shortcuts.} Consider the following illustrative task:
\fcolorbox{black}{gray!6}{\parbox{0.96\linewidth}{
\textbf{Task:} Induce a minimal logic rule for the eastbound trains. It must entail all eastbound trains and no westbound trains, capturing the key underlying relational pattern in the train attributes.

\textbf{B:} \; has\_car(train0,car0). \; car\_color(car0,red). \;
has\_car(train1,car1). \; car\_color(car1,blue). \\
\textbf{E:} \; eastbound(train0). \; westbound(train1).

\smallskip
\hrule
\smallskip

\textbf{Inductive Rule:} eastbound(T) :- has\_car(T,C), car\_color(C,red).

\textbf{Reward Shortcut:} eastbound(train0). \; westbound(train1).
}}\\
Geniune rule induction captures the underlying relational structure of the tasks, producing a logic rule that explains the observed labels, and generalizes to unseen instances. In the example above, a valid inductive rule would be “A train is eastbound if it carries a red car.”
Reward shortcuts, in contrast, bypass rule induction altogether and instead exploit weaknesses in the evaluation protocol. The reward shortcut above correctly assigns the eastbound label to 'train0'; consequently, imperfect verification that only checks for extensional correctness would yield a 'false positive'.

\textbf{Isomorphic Perturbation Testing.}
Detecting shortcut behavior is difficult because correct logic rules do not have a unique syntactic form (logically equivalent rules can differ by literal reordering or variable renaming). Consequently, evaluation often relies on extensional correctness~\cite{Cropper2021InductiveLP}, judging the rule by whether it produces the correct labels on the given examples. Under such evaluation, shortcuts that enumerate examples are indistinguishable from genuine rule induction. IPT resolves this ambiguity by testing invariance under logical isomorphisms.
For each task $\mathcal{T} = (B, E^+, E^-)$, the model produces a single hypothesis $H$, which is evaluated under two verification regimes. (1) \textit{Extensional verification} checks completeness and consistency on the task using the task's object identifiers (e.g., \symbols{train0}, \symbols{car0}). (2) \textit{Isomorphic verification} checks completeness and consistency on a logically isomorphic task $\mathcal{T}^\Phi = (B^\Phi, E^{+\Phi}, E^{-\Phi})$, obtained under a bijective renaming of object constants $\Phi: c \to \Phi(c)$, while attribute constants (e.g., \symbols{red}, \symbols{short}) remain fixed. Applying $\Phi$ to the earlier example yields:
\fcolorbox{black}{gray!6}{\parbox{0.96\linewidth}{
\textbf{Perturbed Example Task (under mapping $\Phi$)}\\
\textbf{B:} has\_car(t1, c1). car\_color(c1, red). has\_car(t2, c2). car\_color(c2, blue).\\
\textbf{E:} eastbound(t1). westbound(t2).
}}
Because the two verification settings are logically isomorphic, any hypothesis that captures the underlying relational structure remains valid under both. In contrast, hypotheses that rely on specific object identifiers (e.g., \symbols{train0}) fail under logically isomorphic verification, as they no longer appear.

\textbf{Quantifying Reward Shortcuts.} Shortcut behavior is identified by comparing outcomes under the two verification regimes. Formally, a hypothesis $H$ is a \textit{reward shortcut} w.r.t.\ task $\mathcal{T}$ and perturbation $\Phi$ if it is complete and consistent on the original task $\mathcal{T}$, but not on its isomorphic perturbation $\mathcal{T}^\Phi$.  This provides a direct, model-agnostic criterion for detecting shortcut reliance from outputs alone.

\section{Monitoring Shortcut Behaviour}\label{sec:results}
\begin{figure}[t]
    \centering
    \begin{subfigure}[t]{.48\textwidth}
        \centering
        \includegraphics[width=\textwidth]{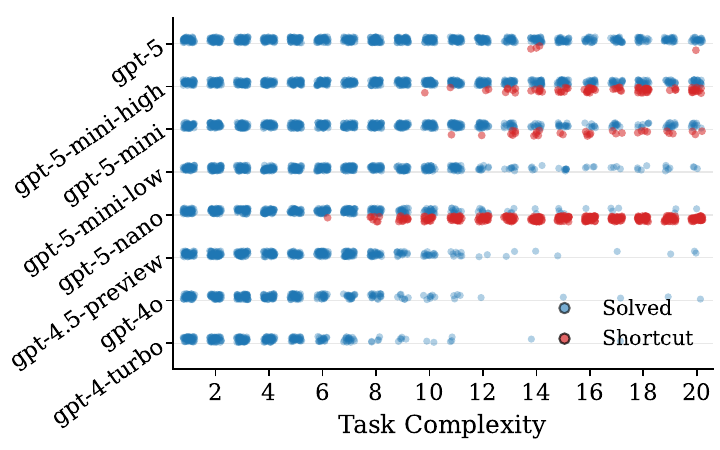}
        \caption{As task complexity increases, RLVR models increasingly resort to shortcut behaviours.}
        \label{fig:shortcut_complexity}
    \end{subfigure}
    \quad 
    \begin{subfigure}[t]{.48\textwidth}
        \centering
        \includegraphics[width=\textwidth]{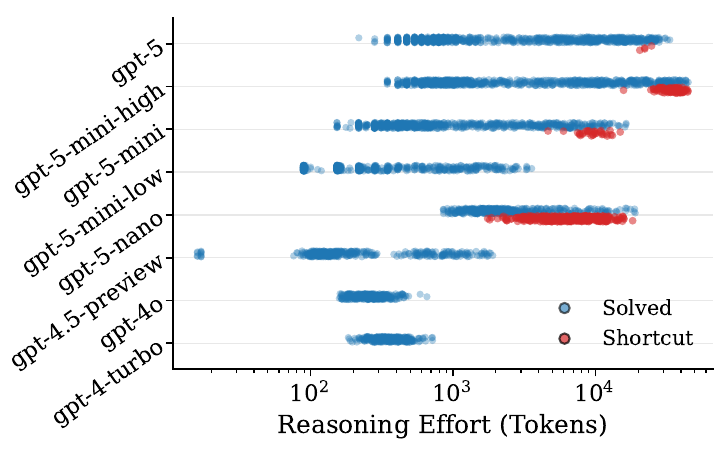}
        \caption{As inference-time compute increases, RLVR models increasingly resort to shortcut behaviors.}
        \label{fig:shortcut_compute}
    \end{subfigure}
    \caption{Shortcut behavior scales with task complexity and inference-time compute. RLVR-trained models exhibit increasing shortcut prevalence as tasks become harder and more compute is allocated.}
    \label{fig:shortcut}
    \vskip -1em
\end{figure}

We evaluate reward shortcut behavior using IPT across frontier models, including RLVR-trained reasoning models (GPT-5, Olmo3), non-RLVR reasoning models (Ministral), and conventional LLMs (GPT-4), on the \benchmark~\citep{helff2025slr} benchmark of logical reasoning tasks with increasing complexity. Each model produces a single output per task, which is evaluated under both extensional and isomorphic verification, enabling us to distinguish genuine rule induction from shortcut strategies. \autoref{tab:shortcut_overview} reports accuracy and shortcut counts, while \autoref{fig:shortcut} provides a per-task diagnostic view of shortcvut behavior. Complementing this, \autoref{fig:shortcut_trends} characterizes aggregate scaling trends, and \autoref{fig:training_dynamics} links shortcut behavior to RLVR training dynamics.

\textbf{RLVR models exhibit systematic shortcut behavior.}
A clear dichotomy emerges between model families. Across our evaluation (see \autoref{fig:shortcut}, \autoref{tab:shortcut_overview}), all non-RLVR models (GPT-4 family, Ministral) exhibit zero shortcuts. In contrast, RLVR-trained models (GPT-5 family, Olmo3) consistently produce reward shortcuts despite stronger benchmark performance. This indicates that shortcut behavior is not an inherent limitation of LLMs, but a failure mode specific to RLVR-based reasoning models.

\textbf{Task Complexity Drives Shortcut Behavior.} \autoref{fig:shortcut_complexity} shows a strong correlation between task difficulty and shortcut behavior. For example, 70\% of the shortcuts produced by \symbols{gpt-5-mini-high} occur in the highest-complexity quartile. Aggregated across all models, only 40 shortcuts appear in the first 10 complexity levels, compared to 458 in levels 11–20. This trend (also reflected in \autoref{app:fig:complexity_trend}) suggests that as the cost of genuine induction increases, models increasingly resort to shortcut strategies.\looseness=-1

\textbf{Inference-Time Compute Drives Shortcut Behavior.} \autoref{fig:shortcut_compute} shows that shortcuts are not uniformly distributed across reasoning effort, but concentrate at high token budgets. Consistently, \autoref{app:fig:gpt5_mini_trend} shows that increasing the reasoning effort of \symbols{gpt-5-mini} from \textit{low} to \textit{medium} and \textit{high} raises shortcut counts from 0 to 32 and 84, respectively. This suggests that additional compute may not be used solely to improve reasoning, but may also be allocated to discovering and exploiting reward shortcuts.

\textbf{Anatomy of a Shortcut.} We observe two recurring shortcut patterns, both of which revert to extensional enumeration strategies.
\textit{1. Blatant Enumeration.} The model abandons the required rule structure and lists positive examples as grounded facts rather than inducing shared relational properties (e.g., car color or payload). This direct extensional collapse appears in GPT-5-mini (Problem 685):
\fcolorbox{black}{gray!6}{\parbox{0.96\linewidth}{
\textbf{Blatant Enumeration:} \quad \quad \;
eastbound(train0). eastbound(train1). ... eastbound(train9).
}}\\
\textit{2. Obfuscated Enumeration.} A more sophisticated variant disguises enumeration within rule syntax by encoding disjunctions over specific object identifiers. GPT-5 exhibits this behavior in Problem 686:
\fcolorbox{black}{gray!6}{\parbox{0.96\linewidth}{
\textbf{Obfuscated Enumeration:} \quad
eastbound(T) :- has\_car(T,car0\_1);  ...; has\_car(T,car10\_1). 
}}
Both forms reflect failures of inductive reasoning, but the obfuscated variant is particularly concerning. It visually mimics valid hypotheses while preserving shortcut behavior. This suggests optimization pressure not only to exploit verifier weaknesses, but also to conceal such exploitation.

\textbf{RLVR can Induce Reward Shortcuts.}
The inference-time results establish a strong association between RLVR and shortcut behavior. To probe causality, we run a controlled training experiment (\autoref{app:sec:training}) in which two identical base models are trained with \cite{olmo2025olmo3} \mbox{RLVR pipeline}, differing only in the verifier used for reward. When trained against the extensional verifier, the model develops a growing \emph{hacking gap}, a divergence between extensional and isomorphic reward that emerges mid-training and continues to widen (see \autoref{fig:training_dynamics}). In contrast, training with an isomorphic verifier keeps this gap near zero throughout. These results show that imperfect extensional verification induces reward shortcut strategies, while isomorphic verification removes this incentive. These findings suggest that such strategies are learned during training and may persist at deployment.

\section{Conclusion}
We identify reward shortcuts as a systematic failure mode in RLVR-trained reasoning models, where models exploit weaknesses in verifiers rather than performing genuine rule induction. With IPT, we provide a black-box method to detect such behaviors in frontier systems without requiring access to weights or reasoning traces. Our findings show that shortcut prevalence increases with both task complexity and inference-time compute, and that such behavior is not merely correlational but can be directly induced by the training signal. These results highlight a critical misalignment risk in RLVR training and motivate evaluation protocols that more faithfully enforce intended reasoning objectives.

\section*{Acknowledgments} We acknowledge support of the hessian.AI Innovation Lab (funded by the Federal Ministry of Research, Technology and Space, BMFTR, grant no. 16IS22091), the hessian.AISC Service Center (funded by the Federal Ministry of Education and Research, BMBF,  grant No 01IS22091), and the Center for European Research in Trusted AI (CERTAIN). Further, this work benefited from the ICT-48 Network of AI Research Excellence Center ``TAILOR'' (EU Horizon 2020, GA No 952215), the Hessian research priority program LOEWE within the project “WhiteBox”, the HMWK cluster projects ``Adaptive Mind'' and ``Third Wave of AI'', and from the NHR4CES. This work has also benefited from the BMWE project "Sovereign Open Source Foundational Models für European Intelligence (SOOFI)," 13IPC040G, and also from early stages of the Cluster of Excellence "Reasonable AI" funded by the German Research Foundation (DFG) under Germany’s Excellence Strategy— EXC-3057; funding will begin in 2026. This work was supported by the Priority Program (SPP) 2422 in the subproject “Optimization of active surface design of high-speed progressive tools using machine and deep learning algorithms“ funded by the German Research Foundation (DFG). Further, this work was funded by the AlephAlpha Collaboration lab 1141. This work was supported in part by OpenAI Research Credits.

\bibliography{bibliography}
\bibliographystyle{iclr2026_conference}

\appendix
\section*{Supplementary Material}
\addcontentsline{toc}{section}{Supplementary Material}

\addcontentsline{toc}{section}{Supplementary Material}

\section{Limitations}
Our analysis is conducted on a single benchmark domain (SLR-Bench), which frames inductive reasoning through logic programming over train classification tasks. While the shortcut behaviors we identify are systematic and reproducible, the extent to which they generalize to other reasoning domains (e.g., mathematical, causal, or abductive reasoning) remains an open question. Furthermore, our evaluation of frontier models (GPT-5 family) is limited to black-box access, preventing direct inspection of reasoning traces or internal representations. IPT detects shortcuts through behavioral invariance testing, but cannot distinguish whether shortcut strategies are explicitly represented in the model's reasoning process or emerge implicitly from output distributions. Finally, our controlled training experiment uses a 7B-parameter model due to computational constraints; whether the observed training dynamics scale identically to larger model sizes warrants further investigation.

\section{Detailed Shortcut Analysis} \label{app:sec:shortcut_analysis}
\begin{figure*}[t]
    \centering
    \begin{subfigure}[t]{.47\textwidth}
        \centering
        \includegraphics[width=\textwidth]{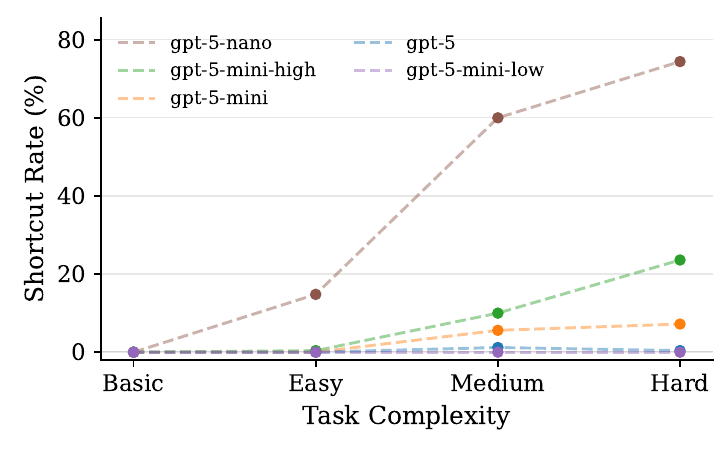}
        \caption{As models face more complex reasoning tasks, they increasingly resort to shortcut behaviours.}
        \label{app:fig:complexity_trend}
    \end{subfigure}
    \hfill
    \begin{subfigure}[t]{.47\textwidth}
        \centering
        \includegraphics[width=\textwidth]{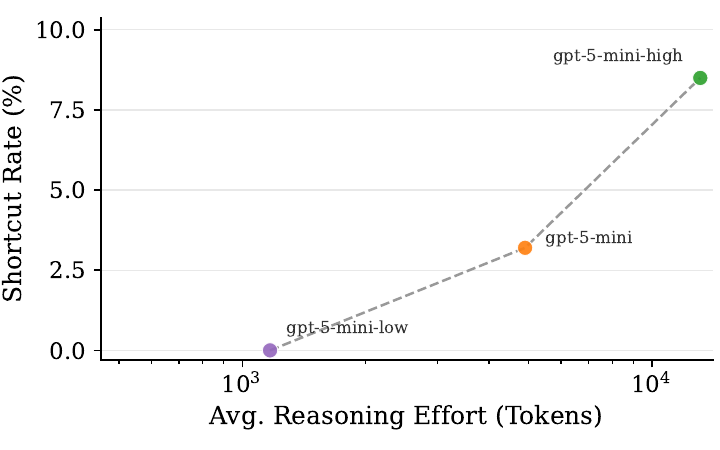}
        \caption{As the reasoning effort of gpt-5-mini is scaled, the model increasingly resorts to shortcut behaviours.}
        \label{app:fig:gpt5_mini_trend}
    \end{subfigure}
    \caption{{Shortcut rate} ($\text{shortcuts}/\text{num tasks}$) as a function of task complexity and inference-time compute. Left: shortcut rate by complexity tiers. Right: shortcut rate by reasoning effort. Trends show that both increasing task difficulty and inference compute drive shortcut prevalence.}
    \label{fig:shortcut_trends}
\end{figure*}

A detailed overview of the entire evaluation is outlined in \autoref{tab:shortcut_overview}, along with aggregated trends in \autoref{fig:shortcut_trends}. The benchmark consists of tasks across four complexity tiers, each consisting of 5 complexity levels: \textit{Basic} (level 1-5), \textit{Easy} (level 6-10), \textit{Medium} (level 11-15), and \textit{Hard} (level 16-20). Each model performs a single inference pass per task, and the resulting hypothesis is evaluated under both extensional and isomorphic verification.

\autoref{tab:shortcut_overview} reports tier-wise accuracy and the number of reward shortcuts ($N_S$). Accuracy is defined as the percentage of tasks solved under isomorphic verification, requiring genuine rule induction. The shortcut count $N_S$ measures the number of tasks (out of $250$ per tier) where a hypothesis satisfies extensional verification but fails under isomorphic verification. In addition, \autoref{fig:shortcut_trends} reports the \textit{shortcut rate}, defined as the ratio of shortcuts to the total number of tasks, i.e.,  $\text{shortcut rate} = N_S / N_{\text{tasks}}$.

\textbf{Model-scale - shortcut trend.}
\autoref{tab:shortcut_overview} reveals substantial variation across model scales. Larger models such as \symbols{gpt-5} exhibit relatively few shortcuts, whereas smaller models (e.g., \symbols{gpt-5-mini-high}, \symbols{gpt-5-nano}) show significantly higher shortcut counts. Notably, \symbols{gpt-5-nano} exhibits extreme shortcut reliance in higher complexity tiers. This suggests that smaller models possess a weaker internal representation of the task objective, making them more susceptible to derailing into shortcut strategies rather than pursuing genuine rule induction. Larger models, by contrast, appear to maintain a more robust understanding of the underlying reasoning structure, resorting to shortcuts primarily as a deliberate fallback when task complexity exceeds their inductive capacity. Then, extensional enumeration offers a viable strategy to game the verifier rather than returning no reward at all.

\textbf{RLVR optimization pressure - shortcut trend.}
\autoref{tab:shortcut_overview} shows that \symbols{olmo-3-32b} exhibits no shortcut behavior. In contrast, \symbols{olmo-3.1} trained under the same setup but with extended RLVR optimization begins to exhibit shortcuts. This indicates that shortcut strategies are not merely present, but are actively \emph{discovered and reinforced} through optimization. As training progresses, RL increasingly amplifies behaviors that maximize the reward signal. When the verifier is imperfect, this creates an optimization landscape in which shortcut strategies can yield high reward without requiring genuine reasoning. Over time, these strategies become more prominent, suggesting that continued optimization pressure can shift the model toward policies that are better at exploiting the verifier rather than solving the underlying task.

\textbf{Task complexity - shortcut trend.}
Across models, shortcut behavior is heavily concentrated in the \textit{Medium} and \textit{Hard} tiers. As shown in \autoref{fig:shortcut_trends} (left), shortcut rates remain low for \textit{Basic} tasks, but increase sharply with complexity. This suggests a qualitative shift in strategy: when tasks are simple, models can satisfy the objective via genuine rule induction; as complexity increases and induction becomes more difficult, optimization pressure favors alternative strategies that achieve high reward at lower cost. Shortcut behavior thus appears not as a random failure, but as a systematic fallback when induction becomes computationally or search-wise expensive.

\textbf{Reasoning effort - shortcut trend.}
\autoref{fig:shortcut_trends} (right) shows that increasing inference-time compute leads to higher shortcut rates. For the \symbols{gpt-5-mini} family, scaling reasoning effort from low to high results in a monotonic increase in shortcut prevalence. This indicates that additional compute is not inherently aligned with better reasoning. Instead, it expands the search over possible strategies, including those that exploit weaknesses in the verifier. In this sense, more compute amplifies the model’s ability to discover reward-maximizing behaviors—whether or not they correspond to the intended reasoning process.

\begin{table*}[t]
\centering
\resizebox{\textwidth}{!}{
\begin{tabular}{lcc | rrrr | rrrr | rrr}
\toprule
& \multicolumn{2}{c}{\textit{Reasoning RL}} 
& \multicolumn{4}{c}{\textit{Reasoning Accuracy (\%)}} 
& \multicolumn{4}{c}{\textit{\# Shortcuts ($N = 250$)}} 
& \multicolumn{3}{c}{\textit{Efficiency \& Cost}} \\
\cmidrule(lr){2-3} \cmidrule(lr){4-7} \cmidrule(lr){8-11} \cmidrule(lr){12-14}
Model & Judge & RLVR
& Basic & Easy & Med. & Hard 
& Basic & Easy & Med. & Hard 
& Syntax & Tokens & USD \\
\midrule

\rowcolor{reasoningllm!50} Gpt-5 & (\cmark) & (\cmark)
& 100 & 100 & 77 & 50 & 0 & 0 & 3 & 1 & 100 & 9.4M & 103.13 \\

\rowcolor{reasoningllm!50} Gpt-5 Mini\textsuperscript{H} & (\cmark) & (\cmark)
& 100 & 100 & 74 & 44 & 0 & 1 & 23 & 59 & 93 & 13.1M & 27.98 \\

\rowcolor{reasoningllm!50} Gpt-5 Mini\textsuperscript{M} & (\cmark) & (\cmark)
& 100 & 98 & 50 & 23 & 0 & 0 & 14 & 18 & 98 & 4.9M & 11.54 \\


\rowcolor{reasoningllm!50} Gpt-5 Mini\textsuperscript{L} & (\cmark) & (\cmark)
& 100 & 85 & 26 & 8 & 0 & 0 & 0 & 0 & 98 & 1.2M & 4.07 \\

\rowcolor{reasoningllm!50} Gpt-5 Nano & (\cmark) & (\cmark)
& 99 & 74 & 12 & 3 & 0 & 37 & 147 & 184 & 99 & 6.2M & 2.81 \\

\midrule

\rowcolor{reasoningllm!50} OLMo-3.1 32B & \cmark & \cmark
& 81 & 60 & 11 & 2 & 2 & 1 & 3 & 7 & 98 & 14.6M & -- \\

\rowcolor{reasoningllm!50} OLMo-3 32B & \cmark & \cmark
& 99 & 68 & 11 & 2 & 0 & 0 & 0 & 0 & 98 & 16.0M & 9.04 \\

\rowcolor{reasoningllm!50} OLMo-3 7B & \cmark & \cmark
& 30 & 15 & 1 & 0 & 0 & 0 & 0 & 0 & 95 & 17.8M & -- \\

\midrule

\rowcolor{reasoningllm!50} Ministral-3 14B & \cmark & \xmark
& 90 & 74 & 17 & 7 & 0 & 0 & 0 & 0 & 50 & 2.7M & 0.82 \\

\rowcolor{reasoningllm!50} Ministral-3 8B & \cmark & \xmark
& 90 & 63 & 10 & 2 & 0 & 0 & 0 & 0 & 47 & 1.5M & 0.43 \\

\rowcolor{reasoningllm!50} Ministral-3 3B & \cmark & \xmark
& 79 & 47 & 7 & 2 & 0 & 0 & 0 & 0 & 61 & 3.5M & 0.77 \\

\midrule

\rowcolor{basellm!50} Gpt-5 (chat) & (\xmark) & (\xmark)
& 100 & 91 & 34 & 14 & 0 & 0 & 0 & 0 & 100 & 2.7M & 36.04 \\

\rowcolor{basellm!50} Gpt-4.5 Preview & (\xmark) & (\xmark)
& 96 & 61 & 6 & 2 & 0 & 0 & 0 & 0 & 100 & 0.4M & 576.40 \\

\rowcolor{basellm!50} Gpt-4o & (\xmark) & (\xmark)
& 95 & 31 & 2 & 1 & 0 & 0 & 0 & 0 & 100 & 0.3M & 20.03 \\

\rowcolor{basellm!50} Gpt-4o-mini & (\xmark) & (\xmark)
& 92 & 18 & 0 & 0 & 0 & 0 & 0 & 0 & 100 & 0.4M & 1.26 \\

\rowcolor{basellm!50} Gpt-4 Turbo & (\xmark) & (\xmark)
& 93 & 20 & 2 & 0 & 0 & 0 & 0 & 0 & 100 & 0.4M & 81.30 \\

\bottomrule
\multicolumn{14}{l}{\small Parenthesized values (\cmark/\xmark) indicate presumed training methodology. Reasoning effort: \textsuperscript{L}: Low, \textsuperscript{M}: Medium, \textsuperscript{H}: High, --: No pricing information available.}
\end{tabular}
}
\caption{Comparison of logical reasoning accuracy, reward shortcuts, and efficiency across models.}
\label{tab:shortcut_overview}
\end{table*}

\begin{figure*}[t]
    \centering
    \begin{subfigure}[t]{.47\textwidth}
        \includegraphics[width=\textwidth]{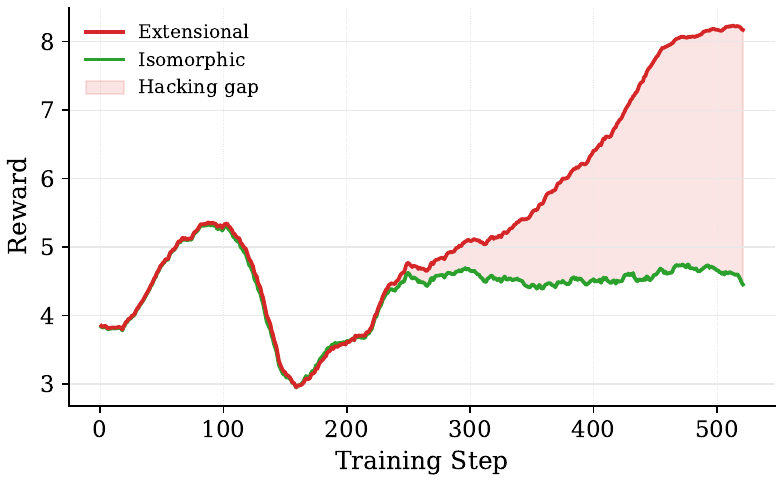}
        \caption{Extensional RLVR: extensional reward diverges as the model learns to exploit the verifier.}
    \end{subfigure}
    \hfill
    \begin{subfigure}[t]{.47\textwidth}
        \includegraphics[width=\textwidth]{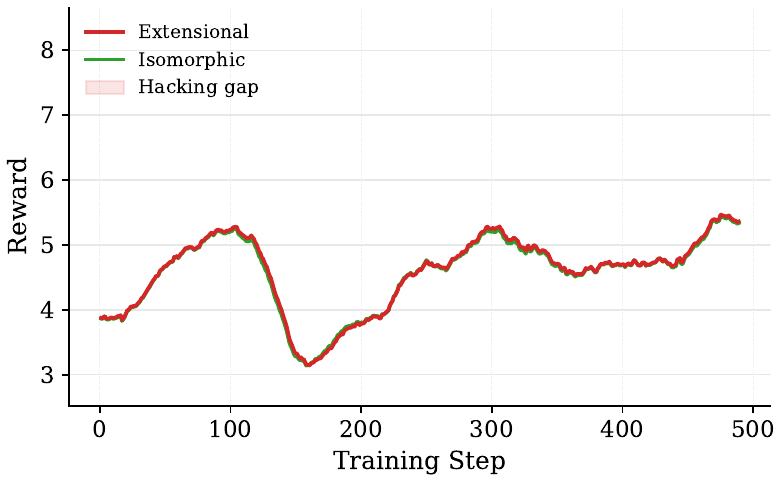}
        \caption{Isomorphic RLVR: both rewards track each other throughout, with no hacking gap.}
    \end{subfigure}
    \caption{Training Olmo-3-7B-Think-DPO via extensional vs. isomorphic RLVR. The hacking gap (shaded) only emerges when the model is trained against an extensional verifier.}
    \label{fig:training_dynamics}
\end{figure*}
\section{RLVR under Isomorphic vs. Non-Isomorphic Reward}\label{app:sec:training}

To validate that the reward shortcuts observed in frontier models can be a consequence of RLVR optimization of extensional verifiers, we conduct a controlled training experiment on \benchmark. We train two variants of the same base model under identical conditions, differing only in the reward signal: one receives feedback from the \emph{extensional verifier}, the other from the \emph{isomorphic verifier}. 

To validate that the reward shortcuts observed in frontier models can arise from RLVR optimization against extensional verification, we conduct a controlled training experiment on \benchmark. We follow the default Olmo-3 RLVR setup (\texttt{Olmo-core} + \texttt{Open Instruct}) \cite{olmo2025olmo3} and finetune two variants of Olmo-3-7B-Think-DPO using \benchmark\cite{helff2025slr}. The two runs only differ in the reward signal: one receives feedback from the \emph{extensional verifier}, the other from the \emph{isomorphic verifier}. We train for about 500 steps per run using 64 H100 GPUs for about 48h.

\autoref{fig:training_dynamics} reports the extensional and isomorphic rewards throughout training. The maximum reward under the Olmo-3 RLVR setup is 10. In the extensional run, both rewards initially track each other; around step 250 they diverge sharply. The extensional reward continues to climb while the isomorphic reward plateaus, indicating that the model has discovered and exploited shortcut strategies that satisfy the extensional verifier without performing genuine rule induction. The shaded region (hacking gap, $r_{\text{ext}} - r_{\text{iso}}$) grows monotonically, reaching a gap of approximately 3.5 reward points after 500 steps. In the isomorphic run, the gap remains near zero throughout, confirming that training against the isomorphic verifier prevents shortcut behaviour. These results provide direct causal evidence that an imperfect extensional reward signal is sufficient to induce reward hacking, and that isomorphic verification removes this incentive.

\end{document}